\documentclass{article} 
\usepackage{colm2024_conference}

\usepackage{booktabs}
\usepackage{graphicx}
\usepackage{enumitem}
\usepackage{wrapfig}
\usepackage{algorithm}
\usepackage{algpseudocode}
\usepackage{natbib}
\usepackage{makecell}
\usepackage{booktabs}
\usepackage{array}
\usepackage{amsmath}
\usepackage{amsfonts}
\usepackage{threeparttable}
\usepackage{multirow}
\usepackage{verbatim}
\usepackage{caption}
\usepackage{longtable}
\usepackage{ragged2e}      
\usepackage{array}         
\usepackage{supertabular}
\usepackage{CJKutf8}
\usepackage{array} 
\usepackage[utf8]{inputenc} 
\usepackage[T1]{fontenc}
\usepackage[french,vietnamese,mongolian,greek,english]{babel}
\usepackage{pifont}
\usepackage{xcolor}
\usepackage{enumitem}
\usepackage{tablefootnote}
\usepackage{xspace}
\usepackage{textcomp}
\usepackage{makecell}
\usepackage{lscape} 
\usepackage{siunitx}
\usepackage{listings}
\usepackage{xcolor}
\usepackage{colortbl}
\definecolor{kuaishoublue}{HTML}{6D9EEB}
\hypersetup{
    colorlinks=true,   
}
\definecolor{dt}{gray}{0.7}
\usepackage{pifont}       
\usepackage{bbding}       
\usepackage{fontawesome}
\newcolumntype{L}[1]{>{\raggedright\arraybackslash}m{#1}}

\usepackage{scrextend}

\usepackage{tgpagella}
\usepackage{latexsym}
\usepackage[T1]{fontenc}
\usepackage[utf8]{inputenc}
\usepackage{microtype}
\definecolor{mydarkblue}{rgb}{0,0.08,0.45}
\definecolor{citecolor}{HTML}{0071BC}
\usepackage{url}            
\usepackage{nicefrac}       
\usepackage{changepage}
\usepackage{xargs}          
\usepackage{wrapfig,lipsum,booktabs}
\usepackage{longtable}
\usepackage{subcaption}
\usepackage{endnotes}

\usepackage{fancyvrb}
\usepackage{fvextra}
\usepackage{pgfplots}
\usetikzlibrary{pgfplots.groupplots}
\pgfplotsset{compat=1.3}
\usepackage{tikz}
\usetikzlibrary{patterns}

\usepackage[most]{tcolorbox}
\usepackage[capitalize,noabbrev]{cleveref}
\crefname{section}{Section}{\S\S}
\Crefname{section}{Section}{\S\S}
\crefname{table}{Table}{Tables}
\crefname{figure}{Figure}{Figures}
\crefname{algorithm}{Algorithm}{}
\crefname{equation}{eq.}{}
\crefname{appendix}{Appendix}{}
\crefformat{section}{Section #2#1#3}
\usepackage{multicol}
\usepackage{tcolorbox}
\newcommand{\abbr}{BaseReward}

\usepackage{titlesec}
\titleformat*{\section}{\large\bfseries}

\definecolor{blue1}{HTML}{196ab1}
\definecolor{blue2}{HTML}{4886c1}
\definecolor{blue3}{HTML}{5e9bd6}
\definecolor{blue4}{HTML}{77b1e2}
\definecolor{blue5}{HTML}{bdd930}
\definecolor{blue6}{HTML}{dfebf6}

\definecolor{red1}{HTML}{de512c}
\definecolor{red2}{HTML}{f2642d}
\definecolor{red3}{HTML}{f68f58}
\definecolor{red4}{HTML}{febf92}
\definecolor{red5}{HTML}{f8e9c8}

\title{BaseReward: A Strong Baseline for Multimodal Reward Model}

\author{
Yi-Fan Zhang$^{*,2}$, Haihua Yang$^{\spadesuit,*,1}$, Huanyu Zhang$^{2}$, Yang Shi$^{4}$\\ Zezhou Chen$^{2}$,
Haochen Tian$^{2}$, Chaoyou Fu$^{3,\dagger}$, Kai Wu$^{1}$, Bo Cui$^{1}$\\  Xu Wang$^{1}$, Jianfei Pan$^{1}$, Haotian Wang$^{5}$,  Zhang Zhang$^{2,\dagger}$, Liang Wang$^{2}$
\\
\footnotesize{
$^{1}$~ByteDance \;
$^{2}$~CASIA \;
$^{3}$~NJU \;
$^{4}$~PKU \;
$^{5}$~THU \;\\
$^{\spadesuit}$~Project Leader \;
$^{*}$~Equal Contribution \;
$^{\dagger}$~Corresponding Author \;}
}

\begin{document}

\maketitle

\begin{abstract}
The rapid advancement of Multimodal Large Language Models (MLLMs) has made aligning them with human preferences a critical challenge. Reward Models (RMs) are a core technology for achieving this goal, but a systematic guide for building state-of-the-art Multimodal Reward Models (MRMs) is currently lacking in both academia and industry. Through exhaustive experimental analysis, this paper aims to provide a clear ``recipe'' for constructing high-performance MRMs. We systematically investigate every crucial component in the MRM development pipeline, including \textit{reward modeling paradigms} (e.g., Naive-RM, Critic-based RM, and Generative RM), \textit{reward head architecture}, \textit{training strategies}, \textit{data curation} (covering over ten multimodal and text-only preference datasets), \textit{backbone model} and \textit{model scale}, and \textit{ensemble methods}.

Based on these experimental insights, we introduce \textbf{BaseReward}, a powerful and efficient baseline for multimodal reward modeling. BaseReward adopts a simple yet effective architecture, built upon a {Qwen2.5-VL} backbone, featuring an optimized two-layer reward head, and is trained on a carefully curated mixture of high-quality multimodal and text-only preference data. Our results show that BaseReward establishes a new \textbf{state-of-the-art (SOTA)} on major benchmarks such as MM-RLHF-Reward Bench, VL-Reward Bench, and Multimodal Reward Bench, outperforming previous open-source and proprietary models. Furthermore, to validate its practical utility beyond static benchmarks, we integrate BaseReward into a real-world reinforcement learning pipeline, successfully enhancing an MLLM's performance across various perception, reasoning, and conversational tasks. This work not only delivers a top-tier MRM but, more importantly, provides the community with a clear, empirically-backed guide for developing robust reward models for the next generation of MLLMs.

\end{abstract}



{
  \setstretch{0.7}
  \tableofcontents
  \noindent\hrulefill
}

\section{Introduction}

The rapid advancement of Multimodal Large Language Models (MLLMs)~\citep{qwen2.5,team2025kwai,zhang2024beyond,coreteam2025mimovltechnicalreport,internvl,fu2025vita} has ushered in a new era of AI capabilities, enabling sophisticated understanding and generation across diverse data modalities, including text, images, video, and audio. Despite these impressive achievements, a central challenge remains: ensuring that these powerful models consistently produce outputs that are helpful, harmless, and aligned with human values and preferences. A pivotal technology to address this challenge is the reward model (RM), which is trained to evaluate and score model outputs based on human feedback. These reward models serve as crucial learning signals for fine-tuning MLLMs via methods such as Reinforcement Learning from Human Feedback (RLHF)~\citep{2023llavarlhf,ouyang2022training,zhang2025mm}, effectively steering the models toward safer, more reliable, and user-aligned behaviors.

While the concept of reward modeling is well-established for text-only Large Language Models (LLMs), the blueprint for constructing state-of-the-art Multimodal Reward Models (MRMs)~\citep{pu2025judge,chen2024mllm,xiong2024llava,wang2025visualprmeffectiveprocessreward,zang2025internlm,zhang2025r1} remains less clear. Currently, state-of-the-art MLLMs, each employ distinct reward modeling strategies, incorporating various domain-specific techniques. For instance, Seed 1.5 VL~\citep{seed2025seed1_5vl} and Keye-VL~\citep{team2025kwai} utilize generative reward models, with the former enhancing reliability by comparing rollout content against golden references. Mimo-VL~\citep{coreteam2025mimovltechnicalreport} employs dual reward models—one specialized for text-only questions and another for multimodal tasks. GLM 4.1 V Thinking~\citep{vteam2025glm45vglm41vthinkingversatilemultimodal} adopts domain-specific reward strategies tailored to different data categories. Despite this diversity in approaches, the research landscape lacks a systematic, comprehensive study to guide researchers effectively. Critical questions remain unanswered: Which reward model architecture delivers optimal performance? What constitutes the most effective architectural design for reward models? How do different data sources—including text-only preference data—influence multimodal performance? What roles do the MLLM backbone architecture and model scale play in determining effectiveness? 

This paper provides a ``recipe'' for building a high-performance MRM by conducting an exhaustive experimental analysis to answer these fundamental questions. We systematically investigate every crucial component of the MRM development pipeline:
\begin{itemize}[leftmargin=*, noitemsep]
    \item \textbf{Reward Modeling Paradigms:} We compare the performance of Naive, Critic-based, and Generative reward models to identify the most efficient and effective approach.
    \item \textbf{Architectural Design:} We perform detailed ablations on the reward head's structure, including the number of layers and the choice of activation functions.
    \item \textbf{Training Strategies:} We analyze the impact of common regularization techniques, such as zero-coefficient regularization and length normalization, on model performance.
    \item \textbf{Data Curation:} We evaluate the influence of over ten different multimodal and text-only preference datasets, revealing the surprising efficacy of text data in enhancing multimodal judgment and the necessity of careful data selection.
    \item \textbf{Backbone and Scale:} We assess how the choice of the underlying MLLM backbone and its parameter scale affect final reward modeling capabilities.
    \item \textbf{Ensemble Methods:} We explore various ensemble strategies to combine the strengths of diverse models, pushing performance beyond what any single model can achieve.
\end{itemize}

Based on insights gained from our extensive experiments, we present \textbf{\abbr{}}, a powerful and efficient baseline for multimodal reward modeling. \abbr{} leverages a simple yet effective architecture built upon the Qwen2.5-VL~\citep{bai2025qwen25vltechnicalreport} backbone, enhanced with an optimized two-layer reward head, and trained on a carefully curated mixture of high-quality multimodal and text-only preference data. Our model sets a new state-of-the-art (SOTA), surpassing previous open-source and proprietary systems, including Claude 3.7 Sonnet and R1-Reward~\citep{zhang2025r1}, across major benchmarks such as MM-RLHF-Reward Bench~\citep{zhang2025mm} (improving by approximately 11\%), VL-Reward Bench~\citep{li2024vlrewardbenchchallengingbenchmarkvisionlanguage} (improving by approximately 18\%), and Multimodal Reward Bench~\citep{yasunaga2025multimodal}. Additionally, to demonstrate its practical utility beyond static benchmarks, we integrate \abbr{} into a real-world reinforcement learning process. As detailed in Section~\ref{sec:rl}, using \abbr{} to provide the reward signal leads to consistent performance gains when fine-tuning an MLLM across a diverse range of perception, reasoning, and conversational tasks. 

\section{Related Work}

\textbf{Multimodal Large Language Models.} The field of MLLMs has seen explosive growth, building on the successes of text-only LLMs to create models with remarkable capabilities in processing and generating blended content~\citep{bai2025qwen25vltechnicalreport, gpt-o1, team2025kwai, zhang2025thyme}. Research has rapidly advanced, with leading models like Qwen2.5-VL~\citep{bai2025qwen25vltechnicalreport}, InternVL~\citep{chen2023internvl, zhu2025internvl3}, and Llama 3-V~\citep{grattafiori2024llama3herdmodels} demonstrating sophisticated understanding of complex visual and textual inputs. Concurrently, the research community is actively tackling key challenges, including extending context length for long-form content~\citep{shen2025longvitascalinglargemultimodal,shi2025mavors}, improving computational efficiency~\citep{zhang2024beyond}, mitigating model hallucinations~\citep{lu2025dama}, and enhancing conversational abilities~\citep{xiong2024llava}. As these models become more powerful, aligning their outputs with human preferences—ensuring they are helpful, harmless, and accurate—has become a paramount challenge. {Reinforcement Learning from Human Feedback (RLHF)} stands out as a cornerstone technique for this alignment process~\citep{ouyang2022training, zhang2025mm,yu2025aligning}. A critical component of RLHF is the {reward model}, which provides the essential learning signal to guide the MLLM towards more desirable behaviors.

\textbf{Multimodal Reward Models.}  The reward models most relevant to this paper are pure text reward models and multi-modal reward models. There are generally three main approaches to reward modeling. The first approach is to directly use a language model or multi-modal model as the reward model by designing precise prompts that allow them to output a score or ranking~\citep{xiong2024llava}. However, this method heavily depends on the model’s instruction-following ability and comprehension. The second approach involves connecting the latent representation of a language model to a reward head (typically an MLP or linear layer), where the model directly outputs a score. During training, the reward modeling is converted into a binary classification task. This approach is computationally efficient, but it lacks interpretability~\citep{liu2024skywork,zang2025internlm,INF-ORM-Llama3.1-70B,lou2024uncertainty,wang2024helpsteer2}. The final type of model simultaneously learns to evaluate the question-answer pair and creates an additional reward head to provide the score~\citep{yu2024self,zhang2025mm}. Despite the proliferation of these methods, the field lacks a systematic study that provides a \textit{fair comparison} across these different paradigms under a unified experimental setup. Furthermore, there has been limited \textit{deep exploration into crucial aspects of reward model architectural design}, such as the optimal structure of the reward head or the impact of different training strategies and data sources. Our work directly addresses these gaps by conducting an exhaustive experimental analysis to establish a clear ``recipe'' for building high-performance MRMs, culminating in our proposed baseline, {\abbr}.

\section{Recipe for Building MRM}
\subsection{Preliminary}

Reward models are a key component for aligning model outputs with human preferences. Typically, a reward model starts with a pretrained  LLM/MLLM $\phi$, where the LLM head $h_l$ is replaced with a linear reward head $l_r$, enabling the model to output a scalar reward value. These models are trained using human-provided pairwise comparisons. Given a query $x$, a preferred response $y_w$ and a less preferred response $y_l$, the reward model is optimized to assign higher rewards to preferred responses:

\begin{equation}
\mathcal{L}_\text{Reward}(\theta) = \mathbb{E}_{x, y_w, y_l} \left[ -\log \sigma\left( r(y_w|x) - r(y_l|x) \right) \right],
\label{equ:reward_loss}
\end{equation}

where $r(y|x)$ is the scalar reward and $\sigma$ is the sigmoid function.

\subsection{Evaluation Benchmarks and Metrics}

We evaluate model performance using both multimodal and text-only reward benchmarks.

\paragraph{Multimodal reward benchmarks.} The multimodal reward benchmarks consist of VL-Reward Bench~\citep{li2024vlrewardbenchchallengingbenchmarkvisionlanguage}, Multimodal RewardBench~\citep{yasunaga2025multimodal}, and MM-RLHF-Reward Bench~\citep{zhang2025mm}. VL-Reward Bench evaluates models using two metrics: \emph{Overall Accuracy}, which measures the proportion of decisions aligning with human preferences, and \emph{Macro Average Accuracy}, which averages accuracy across various task categories to mitigate the effects of task imbalance. Multimodal RewardBench provides a comprehensive evaluation across six key areas: general correctness, preference alignment, knowledge, reasoning, safety, and visual question answering (VQA). It contains 5,000 annotated triplets, each composed of a multimodal prompt along with chosen and rejected responses. The MM-RLHF-Reward Bench uses two metrics: \emph{Traditional Accuracy (Acc)}, which indicates the fraction of cases where the preferred response is correctly identified, and \emph{Acc+}, a stricter metric that requires correct ranking of all response pairs in a sample, emphasizing robustness in challenging cases with subtle ranking differences or hard-to-distinguish pairs.

\paragraph{Text-Only reward benchmarks.} To evaluate the generalization of multimodal reward models to pure text inputs, RMBench and Reward Bench are utilized. RMBench~\citep{liu2024rm} defines three accuracy metrics reflecting difficulty levels: Easy Accuracy assesses the model’s ability to detect differences when style cues are present; Normal Accuracy evaluates performance when responses share the same style; and Hard Accuracy measures the capacity to identify superior responses based solely on content, even when rejected responses have more favorable style. These metrics are computed across four domains—Chat, Safety, Code, and Math. Reward Bench~\citep{lambert2024rewardbench} further evaluates distinct capabilities including basic dialogue quality (Chat), handling of tricky or adversarial questions (Chat Hard), safety in refusal behaviors (Safety), coding and reasoning skills (Reasoning), and consistency with established preference datasets (Prior Sets). Each subtask uses curated prompts and carefully selected chosen/rejected response pairs to test specific aspects of reward modeling.

Because different ablation targets affect various capability dimensions, all benchmarks are evaluated for data ablations to capture comprehensive effects, while architecture ablations generally focus on a subset sufficient to verify performance improvements.

\textbf{Default Training Data and Backbone.}
For our default experimental configuration, we standardize the training data and model backbone to ensure a consistent basis for comparison. We utilize the supervised fine-tuning (SFT) dataset associated with the R1-Reward~\citep{zhang2025r1} model. This dataset comprises approximately 200,000 preference pairs aggregated from established benchmarks, including MM-RLHF~\citep{zhang2025mm}, VL-Feedback~\citep{li2024vlfeedback}, and RLHF-V~\citep{yu2024rlaif}. For the model architecture, we select the Qwen2.5-VL-7B~\citep{bai2025qwen25vltechnicalreport} as our default backbone, providing a strong and representative foundation for our investigations.

\section{Experimental Analysis}

\subsection{Reward Modeling Approaches}\label{sec:3.1}

To establish a strong foundation for our work, we begin by categorizing and evaluating the dominant paradigms in multimodal reward modeling. We identify three principal approaches:

\begin{itemize}[leftmargin=*, noitemsep]
  \renewcommand\labelitemi{$\diamond$}
  \item \textbf{Naive Reward Model (e.g., IXC-2.5-Reward~\citep{zang2025internlm}).} This represents the most direct method, where a linear reward head is placed atop a pretrained MLLM to output a scalar score. While this approach benefits from exceptional speed in both training and inference, it offers limited insight into its decision-making process, thus appearing as a ``black box''.
  
  \item \textbf{Critic-Based Reward Model (e.g., MM-RLHF~\citep{zhang2025mm}).} This paradigm first prompts the model to generate a textual critique or analysis of the response, and then a reward head scores this generated text. This two-step process provides a degree of interpretability and strikes a balance between performance and efficiency. However, its effectiveness is heavily contingent on the quality of the generated critic; a poorly trained critic can act as a bottleneck, degrading overall performance.
  
  \item \textbf{Generative Reward Model (GRM) (e.g., R1-Reward~\citep{zhang2025r1}, Seed-1.5-VL~\citep{seed2025seed1_5vl}).} This approach reframes reward modeling as a generative task. The model directly generates a token or phrase indicating which of two responses is superior. For instance, R1-Reward takes `[Query, Response 1, Response 2]` as input and is trained to output `<think>[reasoning process]</think><answer>[1 or 2]</answer>`, while Seed-1.5-VL simply outputs the text ``1'' or ``2''. GRMs offer strong interpretability and are often more robust against overfitting, but at the cost of significantly higher computational overhead and lower training efficiency.
\end{itemize}

To systematically and fairly compare these paradigms, we benchmark their performance using a standardized training protocol. Each model type is trained on our curated default dataset. For models requiring an SFT phase for reasoning, such as R1-Reward and MM-RLHF, we use GPT-4o to generate the necessary reasoning data. We conduct evaluations on the VL-Reward Bench and Multi-Modal Reward Bench, as they provide fine-grained assessments across critical capabilities like reasoning, mathematics, and safety. The results of this comparison are detailed in Table~\ref{tab:reward_models}, from which we derive several key observations:

\begin{itemize}[leftmargin=*, noitemsep]
  \renewcommand\labelitemi{$\diamond$}
  \item The quality of Critic-RM heavily depends on the quality of reasoning. The original paper uses manually annotated critics and therefore performed slightly better than our implementation, but this approach is hard to scale up.
  \item Seed 1.5 VL's GRM method can achieve a decent reward modeling effect without training (Seed1.5 VL wo training), but shows noticeable improvement after training, indicating that MLLM itself still requires some training to adapt to the reward modeling task.
  \item Long-CoT-GRM shows clear advantages over Naive RM in coding and safety/bias tasks, but in VQA, general, and hallucination tasks, Naive RM generally achieves comparable or even better results.
\end{itemize}

We believe GRM's advantages in safety/coding mainly come from the knowledge intrinsic to MLLM, and Naive-RM is not necessarily worse than GRM after supplementing this training data. Moreover, due to its simplicity and lower computational cost, Naive-RM is easier to apply during reinforcement learning. Therefore, we selecte Naive-RM as the key research focus and comprehensively explored factors influencing Naive-RM.

\begin{table}[]
\caption{\textbf{Comparison of Different Reward Modeling Approaches on Multi-Modal Reward Bench and VL Reward Bench}, evaluating various fine-grained abilities. For a systematic comparison, all models are evaluated using a unified dataset and training strategy to ensure fairness.}
\label{tab:reward_models}
\resizebox{\columnwidth}{!}{%
\begin{tabular}{lccccccccccccc}
\toprule
 &  & \multicolumn{8}{c}{\cellcolor[HTML]{C9DAF8}\textbf{Multi-Modal Reward Bench}} & \multicolumn{4}{c}{\cellcolor[HTML]{FFF2CC}\textbf{VL Reward Bench}} \\
 &  &  & \multicolumn{2}{c}{\textbf{General}} &  & \multicolumn{2}{c}{\textbf{Reasoning}} &  &  &  &  &  &  \\ \cmidrule{4-5}\cmidrule{7-8}
\multirow{-3}{*}{\textbf{Model}} & \multirow{-3}{*}{\textbf{Overall}} & \multirow{-2}{*}{\textbf{Avg}} & \textbf{Correctness} & \textbf{Preference} & \multirow{-2}{*}{\textbf{Knowledge}} & \textbf{Math} & \textbf{Coding} & \multirow{-2}{*}{\textbf{Safety/bias}} & \multirow{-2}{*}{\textbf{VQA}} & \multirow{-2}{*}{\textbf{Avg}} & \multirow{-2}{*}{\textbf{Reasoning}} & \multirow{-2}{*}{\textbf{Hallucination}} & \multirow{-2}{*}{\textbf{General}} \\ \cmidrule{2-14}
Naive-RM & 70.0 & 64.5 & 65.1 & 62.1 & 69.5 & 78.5 & 49.3 & 42.9 & 84.3 & 75.6 & 68.6 & 78.4 & 79.8 \\
Critic-RM (MM-RLHF) & 60.4 & 63.9 & 54.8 & 55.2 & 62.7 & 63.4 & 52.3 & 78.5 & 80.2 & 62.8 & 56.9 & 66.1 & 65.3 \\
GRM (Seed1.5 VL wo Training) & 58.7 & 64.4 & 55.7 & 54.1 & 60.3 & 65.9 & 59.6 & 77.6 & 77.7 & 53.1 & 56.8 & 58.3 & 44.2 \\
GRM (Seed1.5 VL+SFT) & 71.2 & 69.3 & 63.6 & 64.7 & 65.9 & 76.1 & 55.5 & 75.3 & 83.9 & 73.1 & 65.1 & 77.2 & 77.1 \\
LongCoT-GRM (R1-Reward wo RL) & 68.3 & 72.5 & 67.6 & 64.3 & 63.8 & 74.9 & 57.4 & 95.7 & 83.8 & 64.1 & 59.9 & 72.3 & 60.0 \\
LongCoT-GRM (R1-Reward) & 76.8 & 82.2 & 77.5 & 74.0 & 74.9 & 83.1 & 79.6 & 99.6 & 86.5 & 71.4 & 63.8 & 85.7 & 64.8 \\
\bottomrule
\end{tabular}%
}
\end{table}
\begin{table}[]
\caption{\textbf{Comparison of Different Configurations for the Reward Head.} Both the layer number and activation function of the reward head significantly impact the final reward modeling performance.}\label{tab:reward_head_comparison}
\label{tab:reward_head_comparison}
\resizebox{\columnwidth}{!}{%
\begin{tabular}{ccccccccc}
\toprule
\multirow{2}{*}{\textbf{\# Layer}} & \multirow{2}{*}{\textbf{Act Func}} & \multicolumn{5}{c}{\textbf{VL-Reward Bench}} & \multicolumn{2}{c}{\textbf{MM-RLHF-Reward Bench}} \\ \cmidrule{3-7}
 &  & \textit{Reasoning} & \textit{Hallucination} & \textit{General} & \textit{Overall Acc} & \textit{Macro Acc} & \textit{Acc} & \textit{Acc+} \\\midrule
1.0 & None & 64.5 & 67.4 & \textbf{79.1} & 71.2 & 70.3 & 87.1 & 71.1 \\ 
2.0 & None & 66.3 & 68.8 & 77.9 & 71.7 & 71.0 & 90.0 & 71.7 \\
2.0 & Tanh & 64.5 & 76.7 & 78.9 & 74.8 & 73.7 & 90.1 & 76.1 \\\rowcolor{red4!20}
2.0 & Silu & \textbf{67.9} & \textbf{79.7} & \textbf{79.1} & \textbf{76.5} & \textbf{75.6} & \textbf{92.9} & \textbf{80.4} \\
3.0 & Silu & 67.6 & 67.2 & 77.3 & 71.4 & 70.8 & 90.6 & 76.1 \\
4.0 & Silu & 65.4 & 63.4 & 76.9 & 69.1 & 68.6 & 88.2 & 73.9 \\
5.0 & Silu & 66.7 & 73.2 & 78.3 & 73.5 & 72.7 & 88.8 & 73.9 \\
\bottomrule
\end{tabular}%
}
\end{table}
\subsection{Reward Model Design}
Naive reward models typically use a simple linear layer as the reward head. We find that using an MLP for the reward head significantly improves the RM's capability. Two main elements contribute to this:

\begin{itemize}
  \renewcommand\labelitemi{$\diamond$}

    \item \textbf{Layer Number}: The number of layers in the reward model head determines the model's capacity and learning capability. An excessive number of layers can lead to increased training complexity, while too few layers may restrict the model's representational power.
    \item \textbf{Choice of Activation Function}: The activation function is crucial for model training. Different activation functions, such as ReLU or Tanh, affect the model's non-linear mapping ability and the gradient flow during the training process.
\end{itemize}

In Table~\ref{tab:reward_head_comparison}, we summarize the following experimental findings.

\begin{itemize}
  \renewcommand\labelitemi{$\diamond$}

    \item Both the number of layers in the reward head and the choice of activation function have a significant impact on the final performance of the reward modeling. Using only a 1-layer linear head yields the worst results.
    \item The best reward modeling performance is achieved when the number of layers is 2 and the SiLU activation function is used. Other activation functions, as well as more layers, do not bring significant performance gains.
\end{itemize}
In subsequent experiments, we default to using a configuration with 2 layers and SiLU activation function.

\subsection{Training Regularization Strategies}

During the training process of the reward model, we conduct a detailed ablation study on two common regularization strategies~\citep{zhao2024swiftascalablelightweightinfrastructure}.

\begin{itemize}
  \renewcommand\labelitemi{$\diamond$}

    \item \textbf{Zero-Coefficient Regularization.} This technique applies a penalty to encourage the rewards for both chosen ($r_c$) and rejected ($r_r$) responses to be centered around zero. The regularization term is formulated as the mean of the squared sum of the rewards.
    
    \item \textbf{Length Normalization.} This strategy aims to mitigate the reward model's intrinsic bias towards longer responses. It normalizes the predicted reward by the logarithm of the response length.
\end{itemize}

\begin{figure}
    \centering
    \includegraphics[width=\linewidth]{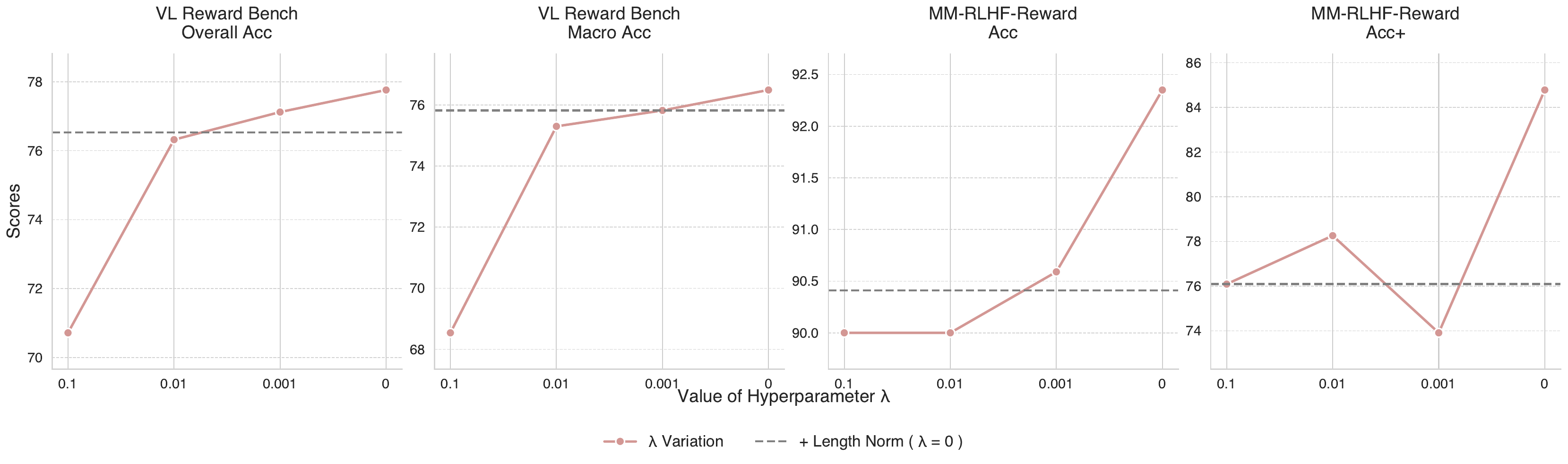}
    \caption{\textbf{The Effect of Different Regularization Strategies on Reward Model Performance.} The solid red line shows the performance variation with $\lambda$. The dashed line represents a baseline model trained with only length normalization and no zero-coefficient regularization ($\lambda=0$). The results show that performance generally declines as $\lambda$ increases from zero.}
    \label{fig:regularization}
\end{figure}

The core ranking loss, which is a function of the reward model's parameters $\theta$, is formally defined as:
\begin{equation}
\mathcal{L}_\text{Reward}(\theta) = \mathbb{E}_{x, y_w, y_l} \left[ -\log \sigma\left( r(y_w|x) - r(y_l|x) \right) \right]
\label{eq:ranking_loss}
\end{equation}
where $\sigma$ is the sigmoid function, $x$ is the prompt, $y_w$ is the preferred (winner) response, and $y_l$ is the rejected (loser) response. The regularization techniques modify these rewards or the overall loss function as described in Algorithm~\ref{alg:regularization}. As illustrated in Figure~\ref{fig:regularization}, we adjust the weight of the zero-coefficient regularizer, $\lambda$, from 0 to 0.1. The results indicate a discernible performance degradation across various metrics as $\lambda$ increases. Furthermore, the inclusion of length normalization alone (represented by the dashed line) does not yield any performance improvement compared to the baseline without regularization (the point where $\lambda=0$). Consequently, we do not apply any regularization loss in the default configuration for training our reward model.

\begin{algorithm}[h!]
\caption{Regularization Strategies for Reward Model Training}
\label{alg:regularization}
\begin{algorithmic}[1]
\State \textbf{Input:} winner rewards $r(y_w|x)$, loser rewards $r(y_l|x)$
\State \textbf{Input:} winner lengths $l_w$, loser lengths $l_l$
\State \textbf{Input:} regularization weight $\lambda$

\Procedure{Length Normalization}{}
    \State $r(y_w|x) \gets r(y_w|x) / \log(l_w + 1.0)$
    \State $r(y_l|x) \gets r(y_l|x) / \log(l_l + 1.0)$
\EndProcedure
\vspace{0.5em}
\Procedure{Loss Computation}{}
    \State $\mathcal{L}_{\text{Reward}} \gets -\text{mean}(\text{logsigmoid}(r(y_w|x) - r(y_l|x)))$
    \State $\mathcal{L}_{\text{zero-coeff}} \gets \lambda \times \text{mean}((r(y_w|x) + r(y_l|x))^2)$
    \State $\mathcal{L}_{\text{total}} \gets \mathcal{L}_{\text{Reward}} + \mathcal{L}_{\text{zero-coeff}}$
    \State \textbf{return} $\mathcal{L}_{\text{total}}$
\EndProcedure
\end{algorithmic}
\end{algorithm}

\begin{table}[t]
    \centering
    \caption{\textbf{Ablation Study Training Datasets.} Dataset size refers to the number of available preference pairs from the original dataset utilized for training the reward model.}
    \label{tab:datasets}
    \begin{tabular}{@{}lllc@{}}
        \toprule
        \textbf{Dataset Name} & \textbf{Type} & \textbf{Size} & \textbf{Source} \\
        \midrule \rowcolor{blue4!20}
        \multicolumn{4}{c}{\textit{Multimodal Preference Data}} \\
        \quad MMIF~\citep{ding2025mm} & Multimodal & 22k & \href{https://huggingface.co/datasets/ChrisDing1105/MMIF-23k}{Link} \\
        \quad Omni-Align~\citep{zhao2025omnialign} & Multimodal & 120k & \href{https://huggingface.co/papers/2502.18411}{Link} \\
        \quad RLAIF-V~\citep{yu2024rlaif} & Multimodal & 83k & \href{https://huggingface.co/datasets/openbmb/RLAIF-V-Dataset}{Link} \\
        \quad MMPR v.12~\citep{zhu2025internvl3} & Multimodal & 2M & \href{https://huggingface.co/datasets/OpenGVLab/MMPR-v1.2}{Link} \\
        \quad R1-Reward~\citep{zhao2025r1} & Multimodal & 200k & \href{https://huggingface.co/datasets/yifanzhang114/MM-RLHF}{Link} \\
        \midrule \rowcolor{red4!20}
        \multicolumn{4}{c}{\textit{Text Preference Data}} \\
        \quad Unltra-All~\citep{cui2023ultrafeedback} & Text-only & 300k & \href{https://huggingface.co/datasets/openbmb/UltraFeedback}{Link} \\
        \quad SHP~\citep{pmlr-v162-ethayarajh22a} & Text-only & 348k & \href{https://huggingface.co/datasets/stanfordnlp/SHP}{Link} \\
        \quad Tulu-3 & Text-only & 65k & \href{https://huggingface.co/datasets/allenai/tulu-3-IF-augmented-on-policy-70b}{Link} \\
        \quad Olmo-2 & Text-only & 378k & \href{https://huggingface.co/datasets/allenai/olmo-2-0425-1b-preference-mix}{Link} \\
        \quad Unltra-Hard & Text-only & 63k & \href{https://huggingface.co/datasets/openbmb/UltraFeedback}{Link} \\
        \quad Others & Text-only & 63k &  \href{https://huggingface.co/datasets/allenai/tulu-3-wildchat-ultrafeedback}{WildChat},  \href{https://huggingface.co/datasets/bigcode/swe-arena-preference-5k}{swe-arena}, etc. \\
        \bottomrule
    \end{tabular}
\end{table}
\subsection{Common Training Datasets}\label{sec:data_select}

\begin{table}[t]
\caption{\textbf{Overall model performance.} Reward models trained on different datasets exhibit significant variation in performance across multimodal and text-only reward benches. Rows highlighted in gray indicate datasets with little or negative performance gains. Ultra-All and Ultra-Hard originate from the same data source but employ different construction strategies; the latter uses only the response pairs with the largest score difference for training. Due to their similar distribution, we retain only the more training-efficient split Ultra-Hard.}

\label{tab:overall_perf}
\resizebox{\columnwidth}{!}{%
\begin{tabular}{lccccc|ccc}
\toprule
\multirow{2}{*}{\textbf{Dateset}} & \multirow{2}{*}{\textbf{\begin{tabular}[c]{@{}c@{}}Multi-Modal\\ Avg\end{tabular}}} & \textbf{VL Reward} & \multicolumn{2}{c}{\textbf{MM-RLHF-Reward}} \textbf{} & \textbf{Multi-Modal Reward} & \multirow{2}{*}{\textbf{\begin{tabular}[c]{@{}c@{}}Pure Text\\ Avg\end{tabular}}} & \textbf{RewardBench} & \textbf{RM Bench} \\ \cmidrule{4-5}
 &   & \textit{Overall} & \textit{Acc} & \textit{Acc+} & \textit{Overall} &  & \textit{Overall} & \textit{Overall} \\ \rowcolor{blue4!20} \midrule
\multicolumn{9}{c}{\textit{Multi-Model Preference Data}} \\ \rowcolor{gray!20} 
MMIF  & 54.3 & 43.2 & 64.9 & 62.4 & 37.0 & 57.6 & 61.2 & 54.0 \\\rowcolor{gray!20} 
Omni-Align  & 49.9 & 46.0 & 61.8 & 30.4 & 30.4 & 60.3 & 66.9 & 53.8 \\
RLAIF-V  & 65.1 & 73.2 & 72.4 & 43.5 & 65.3 & 67.1 & 71.4 & 62.7 \\
MMPR v.12 & 64.0 & 78.7 & 64.1 & 41.3 & 69.8 & 64.7 & 69.9 & 59.4 \\
R1-Reward  & 74.0 & 75.6 & 89.4 & 77.4 & 61.7 & 71.2 & 76.6 & 65.8 \\\midrule \rowcolor{red4!20}
\multicolumn{9}{c}{\textit{Text Preference Data}} \\\rowcolor{gray!20} 
Unltra-All  & 71.7 & 57.1 & 82.3 & 65.2 & 71.1 & 75.3 & 82.1 & 68.5 \\\rowcolor{gray!20} 
SHP  & 54.9 & 35.9 & 68.2 & 39.1 & 55.9 & 61.8 & 66.4 & 57.1 \\
Others   & 68.6 & 65.6 & 84.7 & 63.0 & 71.4 & 65.1 & 73.8 & 56.4 \\
Tulu-3  & 67.8 & 55.1 & 78.8 & 56.6 & 70.1 & 71.2 & 79.9 & 62.6 \\
Olmo-2  & 69.8 & 59.8 & 80.0 & 52.2 & 71.4 & 75.2 & 81.6 & 68.8 \\
Unltra-Hard  & 71.5 & 56.1 & 82.3 & 63.0 & 68.4 & 76.9 & 84.0 & 69.8 \\
\bottomrule
\end{tabular}%
}
\end{table}
In this subsection, we collect over ten datasets, including both multimodal and text-only preference datasets, as detailed in Table~\ref{tab:datasets}. We conduct separate reward model training on each dataset. The final evaluation results are presented in Table~\ref{tab:overall_perf} and Table~\ref{tab:dim_perf}. The former shows the overall performance across all benchmarks, while the latter details the performance for each capability dimension on the VL-Reward Bench and the Multi-Modal Reward Bench. We summarize our experimental findings as follows:

\begin{itemize}[leftmargin=*, noitemsep]
  \renewcommand\labelitemi{$\diamond$}

  \item Certain datasets, such as MMIF and SHP, offer limited benefit to reward model training, likely due to insufficient data diversity or quality issues. Therefore, data curation is essential to avoid introducing unnecessary training overhead or adverse effects.
  
  \item Different datasets influence performance differently. For example, MMPR and RLAIF-V notably enhance results on the hallucination dimension, pushing accuracy on the VL-Reward Bench hallucination metric beyond 90\%. Meanwhile, R1-Reward is particularly effective for reasoning tasks.
  
  \item No single dataset significantly advances reward modeling capability for coding tasks, as reflected by the Multi-Modal Reward Bench results in Table~\ref{tab:dim_perf}. This indicates that specialized downstream tasks require dedicated additional training data.
  
  \item Incorporating text-only data improves multimodal RM performance. For example, training with text-only preference datasets such as Ultra-Hard and Olmo-2 achieves average performance on multimodal benchmarks that is not inferior to multimodal data like MMPR (Multi-Modal Avg in Table~\ref{tab:overall_perf}), and even shows a clear advantage on the Multi-Modal reward bench. This aligns with our hypothesis in Section~\ref{sec:3.1}. As shown in Table~\ref{tab:dim_perf}, the substantial amounts of safety and math content contained in text-only data lead to significant improvements in these dimensions for the reward model, thereby boosting the performance on the Multi-Modal reward bench.
  
  \item To preserve strong text-only reward modeling capability, including text-only datasets in training is necessary. Models trained on virtually any text-only preference data consistently outperform those trained solely on multimodal data in text-based reward benchmarks.
\end{itemize}

\begin{table}[t]
\caption{\textbf{Fine-grained capability analysis.} A detailed analysis of model performance across specific capability dimensions on the VL-Reward and Multi-Modal Reward benchmarks.}
\label{tab:dim_perf}
\resizebox{\columnwidth}{!}{%
\begin{tabular}{lcccccccccccc}
\toprule
\multirow{3}{*}{\textbf{Model}} & \multirow{3}{*}{\textbf{Avg}} &\multicolumn{7}{c}{\textbf{Multi-Modal Reward Bench}} & \multicolumn{4}{c}{\textbf{VL Reward Bench}} \\ \cmidrule{3-9}\cmidrule{11-13}
 &  & \multicolumn{2}{c}{General} & \multirow{2}{*}{Knowledge} & \multicolumn{2}{c}{Reasoning} & \multirow{2}{*}{Safety/bias} & \multirow{2}{*}{VQA} & \multirow{2}{*}{Avg} & \multirow{2}{*}{Reasoning} & \multirow{2}{*}{Hallucination} & \multirow{2}{*}{General} \\ \cmidrule{3-4}\cmidrule{6-7}
 &  & Correctness & Preference &  & Math & Coding &  &  &  &  &  &  \\ \midrule \rowcolor{blue4!20}
\multicolumn{13}{c}{\textit{Multi-Model Preference Data}} \\
RLAIF-V & 65.3 & 61.2 & 49.2 & 61.6 & 67.9 & 46.2 & 81.7 & 79.5 & 73.2 & 55.4 & 92.4 & 71.7 \\
MMPR v.12 & 69.8 & 62.3 & 49.8 & 59.2 & 75.1 & 41.8 & 97.7 & 84.0 & 78.7 & 60.7 & 95.5 & 80.0 \\
R1-Reward & 67.9 & 67.3 & 62.4 & 68.4 & 79.0 & 57.2 & 38.2 & 84.1 & 75.6 & 68.6 & 78.4 & 79.8 \\ \midrule \rowcolor{red4!20}
\multicolumn{13}{c}{\textit{Text Preference Data}} \\
Others & 71.4 & 68.9 & 65.6 & 67.8 & 73.2 & 48.0 & 82.5 & 83.8 & 68.6 & 68.6 & 61.6 & 75.5 \\
Tulu-3 & 70.1 & 62.4 & 61.3 & 64.4 & 72.8 & 44.0 & 94.7 & 83.0 & 61.9 & 61.9 & 60.4 & 63.4 \\
Olmo-2 & 71.4 & 67.6 & 61.5 & 65.4 & 73.9 & 49.0 & 85.4 & 85.6 & 65.0 & 65.0 & 57.6 & 72.5 \\
Unltra-Hard & 68.5 & 63.4 & 57.3 & 66.0 & 76.7 & 53.4 & 60.2 & 85.8 & 59.8 & 59.8 & 62.6 & 56.9 \\
\bottomrule
\end{tabular}%
}
\end{table}

\subsection{Optimizing Multimodal RMs for Pure-Text Tasks}

The preceding analysis establishes the beneficial role of textual data in multimodal reward modeling. This naturally raises a new question: can multimodal preference data, in turn, enhance purely text-based reward modeling tasks? If so, it may be possible to develop a single, comprehensive reward model proficient in both multimodal and text-only domains directly from a multimodal foundation. If not, we must explore alternative strategies to achieve such a versatile RM.

\begin{figure}
    \centering
    \includegraphics[width=\linewidth]{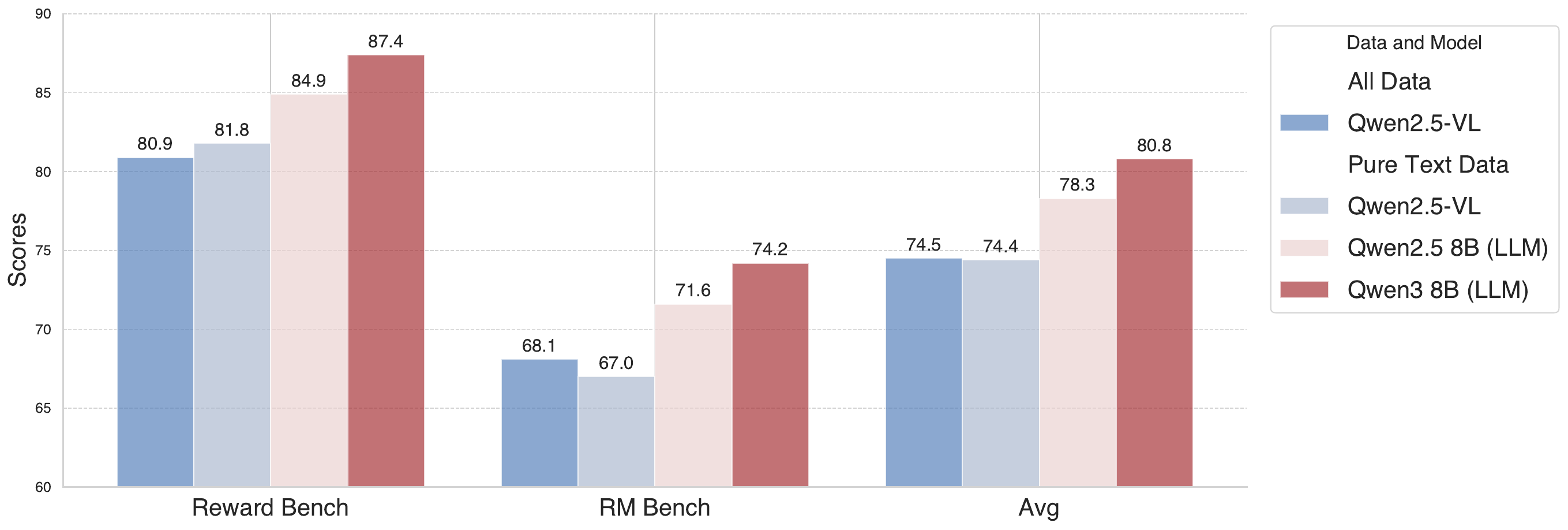}
    \caption{\textbf{Performance Comparison on Pure-text RM Benchmarks.} The MLLM trained with all data (Qwen 2.5 VL-7B) shows no performance gain over the same MLLM trained with text-only data, despite the larger dataset. Both are outperformed by LLMs (Qwen 2.5 8B and Qwen 3 8B) trained on the identical text dataset, highlighting that LLM architectures are more suitable for text-centric reward modeling.}
    \label{fig:pre_text}
\end{figure}

To investigate this, we establish a baseline by training the Qwen 2.5 VL-7B model on seven datasets identified in Section~\ref{sec:data_select} as providing significant gains. For comparison, a second version of this model is trained exclusively on the four text-only datasets from this collection. As Figure~\ref{fig:pre_text} illustrates, the model trained with a larger, mixed-media dataset shows no performance improvement on two pure-text reward model benchmarks, despite the greater volume of data and computational overhead. Furthermore, we train two LLMs, Qwen 2.5 8B and Qwen 3 8B, on the same text-only data. The results indicate that, for a given scale of text data, LLM-based architectures are inherently more adept at pure-text reward modeling than their MLLM counterparts.

Therefore, we conclude that it is not currently optimal to focus on enhancing the multimodal capabilities of a single RM for this purpose. A more effective strategy involves training a dedicated pure-text RM and subsequently integrating it with a multimodal RM. During the reinforcement learning phase, the appropriate RM can be selected dynamically based on the input data type (i.e., text-only or multimodal). This modular approach aligns with methodologies employed in recent studies, such as Mimo-VL.

\subsection{Impact of Base Model Selection and Scale}\label{sec:backbone}

This subsection investigates the influence of different MLLM backbones and their respective scales on final performance. We select a range of prominent models for this analysis, including Intern-VL and Qwen-VL. Our experimental findings are summarized as follows:

\begin{itemize}[leftmargin=*, noitemsep]
  \renewcommand\labelitemi{$\diamond$}

    \item \textbf{Performance Varies Significantly across Model Families and Evaluation Dimensions.} As shown in Table~\ref{tab:backbone}, the Qwen-VL series generally demonstrates superior capability on multimodal reward benchmarks, whereas the Intern-VL series tends to perform better on text-centric benchmarks. For example, on the MM-RLHF-Reward benchmark, Qwen2.5-VL-7B achieves an accuracy of 93.5, which is nearly 10\% higher than the 83.7 achieved by Intern-VL3-8B. Conversely, on RewardBench, Intern-VL3-8B scores 84.0, surpassing the 75.8 score of Qwen2.5-VL-7B. This highlights a clear performance trade-off between different model architectures.

    \item \textbf{Increasing Model Scale Provides Diminishing Returns.} While the size and version of the base model do affect performance, the improvements are not always substantial. The results in Table~\ref{tab:backbone} show that the performance difference between Intern-VL3 at the 2B and 8B scales is marginal across multiple benchmarks. A similar pattern is evident when comparing different versions within the same size class, such as Intern-VL2/3 8B and Qwen2/2.5-VL 7B. This suggests that simply scaling up the MLLM yields limited performance gains. Therefore, for applications with constrained computational resources, models under the 10B parameter scale represent a highly effective and resource-efficient option.
\end{itemize}

\begin{table}[]
\caption{\textbf{Performance Comparison of Various Backbones.} The results highlight the distinct strengths of the Intern-VL and Qwen-VL families across different evaluation criteria. The best performance in each major category is highlighted.}
\label{tab:backbone}
\resizebox{\textwidth}{!}{%
\begin{tabular}{lccccccccc}
\toprule
\multirow{2}{*}{\textbf{Dateset}} & \multirow{2}{*}{\textbf{Scale}} & \multirow{2}{*}{\begin{tabular}[c]{@{}c@{}}\textbf{Multi-Modal}\\ \textit{Avg}\end{tabular}} & \textbf{VL Reward}  &\multicolumn{2}{c}{\textbf{MM-RLHF-Reward}}  & \textbf{Multi-Modal Reward} & \multirow{2}{*}{\begin{tabular}[c]{@{}c@{}}\textbf{Pure Text}\\ \textit{Avg}\end{tabular}} & \textbf{RewardBench} & \textbf{RM Bench }\\
 &  &  & \textit{Overall} & \textit{Acc} & \textit{Acc+} & \textit{Overall} &  & \textit{Overall} & \textit{Overall} \\\midrule \rowcolor{blue4!20}
\multicolumn{10}{c}{\textit{Intern-VL}} \\
Intern-VL2 & 8B & 70.3 & 69.8 & 81.0 & 62.2 & 68.1 & 76.3 & 82.3 & \textbf{70.3} \\
Intern-VL3 & 1B & 62.9 & 67.0 & 77.8 & 54.1 & 52.7 & 65.0 & 68.3 & 61.7 \\
Intern-VL3 & 2B & 71.3 & 73.8 & 83.0 & 62.2 & 66.4 & 70.7 & 75.1 & 66.2 \\
Intern-VL3 & 8B & 72.1 & 74.8 & 83.7 & 62.2 & 67.7 & \textbf{76.8} & \textbf{84.0} & 69.5 \\\midrule \rowcolor{red4!20}
\multicolumn{10}{c}{\textit{Qwen-VL}} \\
Qwen2-VL & 7B & 78.7 & 78.0 & 90.0 & 78.3 & 68.6 & 61.4 & 77.3 & 45.5 \\
Qwen2.5-VL & 3B & 77.9 & 71.1 & 91.8 & 82.6 & 66.2 & 60.8 & 74.9 & 46.7 \\
Qwen2.5-VL & 7B & 80.2 & 79.8 & \textbf{93.5} & \textbf{80.4} & 67.1 & 63.0 & 75.8 & 50.2 \\
Qwen2.5-VL & 32B & \textbf{81.1} & \textbf{82.8} & 92.9 & 78.3 & \textbf{70.5} & 69.1 & 83.4 & 54.8 \\
\bottomrule
\end{tabular}%
}
\end{table}

\subsection{Ensemble Strategies for Reward Models}

In Section~\ref{sec:data_select} and Section~\ref{sec:backbone}, we demonstrate that different data and backbone models exhibit varying impacts across different task dimensions. Consequently, in this subsection, we explore several model ensemble strategies. Our goal is to leverage the complementary strengths of multiple reward models to achieve superior performance simultaneously on both multi-modal and text-only RM tasks. To this end, we utilize the seven datasets selected in Table~\ref{tab:overall_perf} for training. We employ Qwen 2.5 VL 7B and InternVL 3 8B as the backbone models and investigate various ensemble strategies built upon them.

We design several ensemble strategies, which can be categorized as follows. The first category is based on a validation set, for which we uniformly sample 1,000 instances from the seven selected training datasets. For the resulting RMs, we compute normalized weights using three distinct methods:
\begin{itemize}[leftmargin=*, noitemsep]
  \renewcommand\labelitemi{$\diamond$}

    \item \textbf{Accuracy.} The weight is directly determined by the RM's accuracy on the validation set.
    \item \textbf{Confidence.} When an RM evaluates a preference pair (i.e., chosen vs. rejected response), the score margin can be interpreted as its confidence. A larger margin indicates stronger discriminative ability. Therefore, we use the average confidence margin across all validation samples as the weight.
\end{itemize}
In addition to these, we explore a validation-free strategy, which simply involves averaging the reward scores predicted by the individual RMs. The experimental results are presented in Table~\ref{tab:ensemble_results}. Our key observations are as follows:

\begin{itemize}[leftmargin=*, noitemsep]
  \renewcommand\labelitemi{$\diamond$}
    \item \textbf{Significant Performance Gains.} Model ensembling yields substantial improvements on both multi-modal and text-only benchmarks. We observe consistent performance gains across all weighting methods. For instance, on the three multi-modal RM benchmarks, no single model surpasses an average performance of 81.0. However, a simple averaging strategy elevates this score to 82.6.
    \item \textbf{Limited Advantage of Validation-based Methods.} The ensemble strategies based on a validation set require additional data and introduce operational complexity. Despite this, they do not show a clear performance advantage over the simpler averaging strategy.
    \item \textbf{Benefit of Model Diversity.} In the final row of Table~\ref{tab:ensemble_results}, we incorporate an additional model into the ensemble: a Qwen 3 LLM 8B~\citep{qwen3} trained exclusively on the text-only data from our training set. This addition leads to a notable increase in the `Pure Text Avg` performance (from 80.7 to 82.7), demonstrating that enhancing model diversity within the ensemble consistently improves reward modeling capabilities.
\end{itemize}

\begin{table}[]
\caption{\textbf{Performance of Different Ensemble Strategies.} The top section shows the performance of individual reward models. The middle section shows results for ensemble methods that rely on a validation set. The bottom section shows results for validation-free methods.}
\label{tab:ensemble_results}
\resizebox{\columnwidth}{!}{%
\begin{tabular}{lcccc|ccc}
\toprule
\multirow{2}{*}{\textbf{Backbone}} & \multirow{2}{*}{\textbf{\begin{tabular}[c]{@{}c@{}}Multi Modal\\ Avg\end{tabular}}} & \textbf{VL Reward} & \textbf{MM-RLHF} & \textbf{Multi-Modal Reward} & \multirow{2}{*}{\textbf{\begin{tabular}[c]{@{}c@{}}Pure Text\\ Avg\end{tabular}}} & \textbf{Reward Bench} & \textbf{RM Bench} \\
 &  & \textit{Overall} & \textit{Acc} & \textit{Overall} &  & \textit{Overall} & \textit{Overall} \\ \hline
Qwen 2.5 VL 7B & 81.0 & 79.9 & 90.6 & 72.6 & 74.8 & 80.9 & 68.7 \\
InternVL 3 8B & 78.1 & 79.9 & 87.8 & 66.7 & 81.1 & 86.0 & 76.2 \\ 
\midrule \rowcolor{blue4!20}
\multicolumn{8}{c}{\textit{Ensemble Based on Validation Set Performance}} \\
Accuracy & 81.2 & 81.4 & 91.2 & 71.0 & 77.6 & 82.3 & 72.9 \\
Confidence & 80.4 & 81.4 & 88.8 & 71.0 & 77.7 & 82.3 & 73.0 \\
\midrule \rowcolor{red4!20}
\multicolumn{8}{c}{\textit{Validation Set  Free}} \\
Avg & 82.6 & 83.4 & 92.9 & 71.5 & 80.7 & 85.8 & 75.7 \\
+ Qwen 3 LLM 8 B & 82.6 & 83.4 & 92.9 & 71.5 & 82.7 & 88.3 & 77.1 \\
\bottomrule
\end{tabular}%
}
\end{table}

\section{BaseReward}

\subsection{Structure and Training Strategy}

Based on the ablation studies, we propose \abbr, which focuses on multimodal reward modeling. It employs \emph{Qwen2.5-VL-7B} as the backbone and initializes a two-layer MLP as the reward head. The two MLP layers utilize the SiLU activation function between them. The loss function follows Equation~\eqref{equ:reward_loss} without the addition of any auxiliary losses. The training data comprises seven datasets from Table~\ref{tab:overall_perf} that are not marked in gray, aggregating to a total of $2.8$M preference pairs. For the training strategy, a grid search over learning rates $\{1\mathrm{e}{-5}, 3\mathrm{e}{-6}, 1\mathrm{e}{-6}, 3\mathrm{e}{-7}\}$ is conducted, with the final choice of $3\mathrm{e}{-6}$. The batch size is set to $128$, and all training runs complete on $64$ Nvidia H100 GPUs.

Additionally, using the same data and training strategy, we train an extra model adopting \emph{Qwen2-VL-7B} as the backbone, which serves specifically for voting purposes.

\subsection{Baseline Algorithms}
We select several prominent and widely recognized SOTA multimodal models, including GPT-4o-mini (2024-07-18), Claude-3.5-Sonnet (2024-06-22),  Claude-3.7-Sonnet, Gemini-1.5-Flash (2024-09-24), GPT-4o (2024-08-06), Gemini-1.5-Pro (2024-09-24), Gemini-2.0-Flash-Exp, SliME~\citep{zhang2024benchmarking}, VITA-1.5~\citep{fu2025vita}, LLaVA-OneVision-7B-ov~\citep{li2024llava}, Qwen2-VL-7B~\citep{wang2024qwen2}, Molmo-7B~\citep{deitke2024molmo}, InternVL2/3-8B~\citep{chen2023internvl,zhu2025internvl3}, Llama-3.2-11B~\citep{INF-ORM-Llama3.1-70B}, Pixtral-12B~\citep{agrawal2024pixtral}, Molmo-72B~\citep{deitke2024molmo}, Qwen2-VL-72B~\citep{wang2024qwen2} and NVLM-D-72B~\citep{dai2024nvlm}. Furthermore, we compare several recent multimodal reward models, such as \emph{LLaVA-Critic-8B}~\citep{xiong2024llava}, \emph{MM-RLHF-Reward-7B}~\citep{zhang2025mm} and \emph{IXC-2.5-Reward}IXC-2.5-Reward~\citep{zang2025internlm}, which stand at the forefront of recent progress in multimodal reward modeling. The \emph{MM-RLHF-Reward-7B} model operates as a critic-based reward model that first produces an analysis and subsequently utilizes a reward head for scoring. In contrast, \emph{IXC-2.5-Reward} is a classical reward model that directly uses a reward head to score input query-response pairs, achieving state-of-the-art performance across multiple reward benchmarks.

\subsection{Evaluation Results on MRM Benchmark}

The results on \abbr, RLHF-Reward Bench, VL-Reward Bench, and Multi-Modal Reward Bench appear in Tables~\ref{tab:tab_mmrlhf_reward}, \ref{tab:vl_rewardbench}, and \ref{tab:tab_mm_reward}, respectively. Our model, \abbr, surpasses the previous SOTA on MM-RLHF-Reward Bench by $11.9\%$ in accuracy. On the more challenging metric, Acc+, \abbr\ achieves a $23.32\%$ improvement over the prior SOTA \emph{Claude 3.7 Sonnet}. On the VL Reward Bench Overall Accuracy, \abbr\ improves upon the previous best by $14.2\%$. 

It is noteworthy that \abbr\ is a classical reward model featuring very fast inference speed, whereas R1-Reward and MM-RLHF-Reward require an initial critic output step, leading to significantly greater computational overhead. Finally, on the Multi-Modal Reward Bench, \abbr\ achieves the second-best result. This outcome primarily arises from the absence of coding and related preference data in our training set. Additionally, R1-Reward exhibits high sensitivity to prompt design and the ordering of two responses, which increases computational complexity. Section~\ref{sec:rl} details the performance gap between R1-Reward and \abbr\ when applied in the reinforcement learning stage.


\begin{table}[]
\caption{\textbf{MM-RLHF-Reward Bench. }Performance comparison of our reward model (\abbr) with existing open-source and proprietary counterparts.}
\label{tab:tab_mmrlhf_reward}
\resizebox{\textwidth}{!}{%
\begin{tabular}{@{}lcccccccc@{}}
\toprule
\textbf{Models} & \textbf{\#Param} & \textbf{Mcq} & \textbf{Long} & \textbf{Short} & \textbf{Safety} & \textbf{Video} & \textbf{Acc} & \textbf{Acc+} \\ \midrule\rowcolor[HTML]{F5FFFA}
\multicolumn{9}{c}{\texttt{Proprietary Models}} \\ 
Gemini-2.0-Flash-Exp & - & 33.33 & 45.94 & 67.64 & 43.75 & 32.00 & 44.71 & 13.04 \\
GPT-4o (2024-08-06) & - & {64.28} & 78.37 & 44.11 & 56.25 & 40.00 & 58.23 & 26.01 \\
Claude-3.5-Sonnet (2024-06-22) & - &  {64.28} & 67.56 & 55.88 & 65.62 & 60.00 & 62.94 & 26.11 \\
Claude-3.7-Sonnet & - & 66.67 & {91.89} & {91.18} & \underline{87.50} & 76.00 & {82.35} & {65.22}\\
\hline\rowcolor[HTML]{F5FFFA}
\multicolumn{9}{c}{\texttt{Open-Source Models}} \\ 
SliME~\citep{zhang2024benchmarking} & 8B & 23.81 & 10.81 & 14.71 & 12.50 & 7.52 & 17.10 & 1.76 \\
VITA-1.5~\citep{fu2025vita} & 7B & 24.97 & 21.62 & 11.76 & 18.75 & 12.40 & 20.58 & 2.78 \\
Intern-VL-3~\citep{zhu2025internvl3} & 8B & 35.71 & 56.76 & 23.53 & 37.50 & 32.00 & 37.65 & 6.52 \\
NVLM-D~\citep{dai2024nvlm} & 72B & 42.85 & 32.43 & 8.82 & 50.00 & 40.00 & 34.70 & 6.52 \\
Llama-3.2~\citep{INF-ORM-Llama3.1-70B} & 90B & 19.04 & 35.13 & 38.23 & 50.00 & 40.00 & 35.29 & 10.86 \\
Qwen2-VL~\citep{wang2024qwen2} & 72B & 45.23 & 62.16 & 47.05 & 46.88 & 36.00 & 48.23 & 13.04 \\
\hline\rowcolor[HTML]{F5FFFA}\multicolumn{9}{c}{\texttt{\texttt{Reward Models}}} \\
IXC-2.5-Reward~\citep{zang2025internlm} & 7B & 52.38 &  {91.89} & 67.65 & 62.50 & {88.00} & 71.18 & {50.00} \\
MM-RLHF-Reward~\citep{zhang2025mm} & 7B & {83.00} & {97.00} & {74.00} & {69.00} & {88.00} & {82.00} & {63.00} \\
{R1-Reward~\citep{zhao2025r1}} & 7B & {80.95} & {89.19} & {82.35} & {75.00} & {72.00} & {80.59} & {54.35} \\ 
\hline\rowcolor[HTML]{F5FFFA}
\multicolumn{9}{c}{\texttt{Ours}}  \\
\abbr~(Qwen 2 VL) & 7B & {80.95} & \textbf{100.00} & {88.24} & \textbf{90.62} & \textbf{96.00} & {90.59}& \underline{78.26}\\
\abbr~(Qwen 2.5 VL) & 7B & \textbf{95.74} & \underline{97.38} & \underline{94.13} & {81.25} & {88.00} & \underline{91.76}& \textbf{80.43}\\
\abbr~(Ensemble) & 7B+7B & \underline{88.10} & \textbf{100.00} &  \textbf{97.06} & \underline{87.50} & \underline{92.00}& \textbf{92.94} & \textbf{80.43}\\

\bottomrule
\end{tabular}%
}

\end{table}

\definecolor{front-color}{HTML}{FDEFF5}
\begin{table}[]
\centering
\caption{\textbf{VLReward Bench.} Performance comparison of our reward model (\abbr) with existing open-source and private counterparts.}
\label{tab:vl_rewardbench}
\resizebox{\textwidth}{!}{%
\begin{tabular}{@{}lcccccc@{}}
\toprule
\textbf{Models} & \textbf{\#Param} & \textbf{General} & \textbf{Hallucination} & \textbf{Reasoning} & \textbf{Overall Acc} & \textbf{Macro Acc} \\ \midrule \rowcolor[HTML]{FDEFF5} 
\multicolumn{7}{c}{\texttt{Proprietary Models}} \\
Claude-3.5-Sonnet (2024-06-22) & - & 43.40 & 55.00 & 62.30 & 55.30 & 53.60 \\
GPT-4o (2024-08-06) & - & 49.10 & 67.60 & {70.50} & 65.80 & 62.40 \\
Gemini-1.5-Pro (2024-09-24) & - & 50.80 & {72.50} & 64.20 & {67.20} & 62.50 \\
Claude-3.7-Sonnet & - & {68.08} & 70.70 & 60.81 & 66.31 & 66.53 \\
\hline \rowcolor[HTML]{FDEFF5} 
\multicolumn{7}{c}{\texttt{\texttt{Open-Source Models}}} \\
VITA-1.5~\citep{fu2025vita} & 7B & 18.55 & 8.93 & 22.11 & 16.48 & 16.53 \\
SliME~\citep{zhang2024benchmarking} & 7B & 7.23 & 27.09 & 18.60 & 19.04 & 17.64 \\
InternVL2~\citep{chen2023internvl} & 8B & 35.60 & 41.10 & 59.00 & 44.50 & 45.20 \\
LLaVA-Critic~\citep{xiong2024llava} & 8B & 54.60 & 38.30 & 59.10 & 41.20 & 44.00 \\
Molmo~\citep{deitke2024molmo} & 72B & 33.90 & 42.30 & 54.90 & 44.10 & 43.70 \\
Qwen2-VL~\citep{wang2024qwen2} & 72B & 38.10 & 32.80 & 58.00 & 39.50 & 43.00 \\
NVLM-D~\citep{dai2024nvlm} & 72B & 38.90 & 31.60 & 62.00 & 40.10 & 44.10 \\ 
Llama-3.2~\citep{INF-ORM-Llama3.1-70B} & 90B & 42.60 & 57.30 & 61.70 & 56.20 & 53.90 \\
\hline
\rowcolor[HTML]{FDEFF5} \multicolumn{7}{c}{\texttt{\texttt{Reward Models}}} \\
MM-RLHF-Reward~\citep{zhang2025mm} & 7B & 45.04 & 50.45 & 57.55 & 50.15 & 51.01 \\
IXC-2.5-Reward~\citep{zang2025internlm} & 7B & \textbf{84.70} & 62.50 & 62.90 & 65.80 & {70.00 }\\ 
R1-Reward~\citep{zhao2025r1} & 7B & {63.84} & {85.71} & {64.78} & {71.92 }& {71.44} \\
\hline\rowcolor[HTML]{FDEFF5} 
\multicolumn{7}{c}{Ours} \\
\abbr~(Qwen 2 VL) & 7B & {62.12} &{84.82} & \underline{82.64} & {78.53} & {76.53} \\
\abbr~(Qwen 2.5 VL) & 7B & {68.55} &\textbf{92.19} & {81.82} & \underline{82.16} & \underline{80.85} \\
\abbr~(Ensemble) & 7B + 7B & \underline{71.67} &\underline{91.74} & \textbf{85.33} & \textbf{84.41} & \textbf{82.91}
\\ \bottomrule
\end{tabular}%
}
\vspace{-0.3cm}
\end{table}

\begin{table}[t]
\caption{\textbf{Multimodal Reward Bench. }Performance comparison of our reward model (\abbr) with existing open-source and proprietary counterparts.}
\label{tab:tab_mm_reward}
\resizebox{\textwidth}{!}{%
\begin{tabular}{lccccccccc}
\toprule
\multirow{2}{*}{\textbf{Model}} & \multirow{2}{*}{\#\textbf{Param}} & \multirow{2}{*}{\textbf{Overall}} & \multicolumn{2}{c}{\textbf{General}} & \multirow{2}{*}{\textbf{Knowledge}} & \multicolumn{2}{c}{\textbf{Reasoning}} & \multirow{2}{*}{\textbf{Safety}} & \multirow{2}{*}{\textbf{VQA}} \\ \cmidrule{4-5}\cmidrule{7-8}
 &  &  & \textbf{Correctness} & \textbf{Preference} &  & \textbf{Math} & \textbf{Coding} &  &  \\ \hline \rowcolor[HTML]{D9D2E9}
\multicolumn{10}{c}{\texttt{Proprietary Models}} \\
GPT-4o & - & 70.8 & 62.6 & \underline{69.0} & 72.0 & 67.6 & 62.1 & 74.8 & \textbf{87.2} \\
Gemini 1.5 Pro & - & {71.9} & {63.5} & 67.7 & 66.3 & {68.9} & 55.5 & \underline{94.5} & \textbf{87.2} \\
Claude 3.5 Sonnet & - & 71.5 & 62.6 & 67.8 & {73.9} & 68.6 & {65.1} & 76.8 & 85.6 \\  
Claude 3.7 Sonnet &  & {71.9} & 58.4 & 60.7 & \textbf{78.1} & {76.3} & \underline{71.3} & 72.0 & \underline{86.8}  \\
\rowcolor[HTML]{D9D2E9}\hline
\multicolumn{10}{c}{\texttt{Open-Source Models}} \\
SliME~\citep{zhang2024benchmarking} & 8B & 42.0 & 42.3 & 52.2 & 47.5 & 43.5 & 35.3 & 19.1 & 53.8 \\
VITA-1.5~\citep{fu2025vita} & 7B & 53.6 & 55.6 & 54.3 & 52.5 & 51.9 & 52.8 & 58.1 & 50.0 \\
Llama-3.2-Vision-Instruct~\citep{INF-ORM-Llama3.1-70B} & 11B & 51.2 & 57.8 & 65.8 & 55.5 & 50.6 & 51.7 & 20.9 & 55.8 \\
Molmo-D-0924~\citep{deitke2024molmo} & 7B & 52.9 & 56.8 & 59.4 & 54.6 & 50.7 & 53.4 & 34.8 & 60.3 \\
Llama-3.2~\citep{INF-ORM-Llama3.1-70B} & 90B & 61.2 & 60.0 & 68.4 & 61.2 & 56.3 & 53.1 & 52.0 & 77.1 \\
InternVL-3~\citep{zhu2025internvl3} & 8B & 63.6 & 59.6 & 61.6 & 60.5 & 65.1 & 56.6 & 59.3 & 82.3 \\
Qwen-2-VL~\citep{wang2024qwen2} & 72B & 70.9 & 56.4 & 62.3 & 70.2 & 73.3 & 58.9 & 90.1 & 85.3\\  \rowcolor[HTML]{D9D2E9} \hline
\multicolumn{10}{c}{\texttt{Reward Models}} \\
MM-RLHF-Reward~\citep{zhang2025mm} & 7B & 67.1 & 61.7 & 67.5 & 54.3 & 58.4 & 57.9 & 92.9 & 76.8 \\ 
IXC-2.5-Reward~\citep{zang2025internlm} & 7B & 66.6 & 60.7 & 64.2 & 56.8 & 63.0 & 50.5 & 89.9 & 81.1 \\
R1-Reward~\citep{zhao2025r1} & 7B & \textbf{82.2} & \textbf{77.5} & \textbf{74.0} & \underline{74.9} & \textbf{83.1} & \underline{79.6} & \textbf{99.6} & {86.5} \\ 
 \rowcolor[HTML]{D9D2E9} \hline
\multicolumn{10}{c}{\texttt{Ours}} \\
\abbr~(Qwen 2 VL) & 7B & {68.7} & {68.2} & {56.3} & 64.9 & {73.1} & {48.6} & {72.4} & {83.5} \\
\abbr~(Qwen 2.5 VL) & 7B & {72.8} & {65.7} & {65.0} & 70.6 & {82.7} & {50.3} & {81.5} & {85.0} \\
\abbr(Ensemble) & 7B+7B & \underline{73.6} & \underline{68.5} & {68.0} & 70.3 & \underline{82.8} & {51.2} & {81.3} & {85.6} \\
\bottomrule
\end{tabular}%
}
\end{table}


\subsection{Reinforcement Learning with \abbr}\label{sec:rl}

To validate the efficacy of \abbr{} as a reward model, we integrate it into a reinforcement learning pipeline. The ultimate objective of a reward model is to provide high-quality signals for reinforcement learning algorithms. This section examines the performance enhancements achievable by applying \abbr{} in a genuine RL process. Due to computational constraints, we employ a single \abbr{} model (derived from Qwen 2.5 VL) and do not implement a voting or ensemble strategy.

\subsubsection{Experimental Setup}

\textbf{RL Data Curation.}
We curate a diverse dataset for reinforcement learning from a range of prompt sources, including V*~\citep{wu2024v}, arXivQA~\citep{li2024multimodal}, and ThinkLite-VL~\citep{wang2025sota}. These sources respectively target perception, chart recognition, and reasoning tasks. The availability of ground-truth answers in these datasets allows for a comparative study of different reward schemes: a purely rule-based reward, a reward model-based approach, and a hybrid system combining both.

\textbf{Baselines and Training Protocol.}
We employ the Group Relative Policy Optimization (GRPO) \citep{deepseekmath} algorithm to train Qwen-2.5-VL 3B. For each prompt, the process generates 8 rollouts. The training proceeds for one epoch with a batch size of 256. Our primary baseline for comparison is the R1-Reward model, which is the top-performing publicly available general reward model on the MRM benchmark, second only to our own model.

\textbf{Reward Schemes.}
We investigate three distinct reward formulations:

\begin{itemize}
  \renewcommand\labelitemi{$\diamond$}
    \item \textbf{Rule-Based Reward.} This is a binary reward scheme. The reward is 1 if the model's output exactly matches the ground truth and 0 otherwise.
    
    \item \textbf{\abbr-Based Reward.} The reward is directly determined by the score assigned by the \abbr{} model to each response.
    
    \item \textbf{Hybrid Rule-Based + \abbr Reward.} This approach first checks for an exact match with the ground truth. If a match exists, the response receives a reward of 1. Otherwise, the reward is generated by the \abbr{} model and normalized to the range [0, 1] using a sigmoid function. This can be formally expressed as:
    $$ R_{\text{hybrid}}(y) = 
    \begin{cases} 
    1 & \text{if } y \text{ matches ground truth} \\
    \sigma(\text{\abbr{}}(y)) & \text{otherwise}
    \end{cases} $$
    where $y$ is the model response and $\sigma$ is the sigmoid function.
\end{itemize}

For the R1-Reward baseline, which operates on a pairwise preference scoring mechanism, we adopt the following strategy. For the 8 generated responses $\{y_1, \dots, y_8\}$ for a given prompt:
\begin{itemize}[leftmargin=*, noitemsep]
    \item Form all 56 ($8 \times 7$) ordered pairs $(y_i, y_j)$ where $i \neq j$.
    \item For each pair, R1-Reward generates a relative preference score, which we denote as $S(y_i, y_j)$.
    \item The final reward for a response $y_i$ is the aggregation of its preference scores against all other responses:
    $$ R_{\text{R1}}(y_i) = \sum_{j \neq i} S(y_i, y_j) $$
    This score quantifies the collective preference for response $y_i$ over the other candidates.
\end{itemize}

\textbf{Evaluation Benchmarks.}
We assess the performance of the MLLM trained with different reward schemes on a comprehensive suite of benchmarks: MMbench v1.1~\citep{liu2024mmbench}, MME-RealWorld-Lite~\citep{zhang2025mmerealworldmultimodalllmchallenge}, MMStar~\citep{chen2024we}, Mathvista~\citep{lu2024mathvista}, V*~\citep{wu2024v}, Llavawild~\citep{liu2023visual}, and Wildvision~\citep{lu2024wildvision}. These benchmarks are selected to cover a wide array of capabilities: MMbench v1.1 and MMStar function as general-purpose benchmarks; MME-RealWorld-Lite and V* target perceptual abilities; Mathvista focuses on mathematical reasoning; and Llavawild and Wildvision are conversation-oriented benchmarks for holistic evaluation.

\subsubsection{Results and Analysis}

The evaluation results, as detailed in Table~\ref{tab:rl_result}, demonstrate the comparative advantages of our proposed reward strategy. \abbr{} is superior to R1-Reward across all the benchmarks. Furthermore, R1-Reward imposes a significant computational overhead; a substantial portion of the training time is spent awaiting reward generation, leading to suboptimal computational efficiency. A purely rule-based reward mechanism shows marked improvements on the Mathvista benchmark. This is attributable to the objective nature of mathematical problems, where answers are unequivocally right or wrong, making them highly suitable for a binary rule-based system. However, for conversational benchmarks (Llavawild, Wildvision) and general VQA tasks, exclusive reliance on rule-based rewards yields limited performance enhancements, as these tasks often involve nuance and subjectivity that binary rules cannot capture.

The optimal strategy emerges as the hybrid approach combining rule-based checks with \abbr{} scoring. As shown in Table~\ref{tab:rl_result}, this method achieves consistent performance gains across logical reasoning, perception, and conversational tasks. This indicates that the hybrid model effectively leverages the precision of rule-based rewards for objective tasks while utilizing the nuanced, semantic understanding of \abbr{} for more complex and subjective evaluations.

\begin{table}[]
\caption{\textbf{Performance Comparison of the MLLM Trained with Different Rewards}. The hybrid Rule-Based + \abbr approach consistently delivers the most significant improvements.}
\label{tab:rl_result}
\resizebox{\columnwidth}{!}{%
\begin{tabular}{lccccccccc}
\toprule
\multirow{2}{*}{\textbf{Model}} & \textbf{Hallucination} & \textbf{MMbench v1.1} & \multicolumn{2}{c}{\textbf{MME-RealWorld}} & \textbf{MMStar} & \textbf{Vstar} & \textbf{LLaVA-Wild} & \textbf{WildVision} & \textbf{MathVista} \\
 & \textit{Overall} & \textit{Overall} & \textit{Perception} & \textit{Reasoning} & \textit{Overall} & \textit{Overall} & \textit{Score} & \textit{Win Rate} & \textit{Acc} \\\midrule \rowcolor{blue4!20}
\multicolumn{10}{c}{Baseline} \\
Qwen-VL-3B & 43.1 & 77.7 & 45.2 & 36.9 & 54.7 & 74.9 & 82.3 & 48.4 & 61.8 \\
R1-Reward & 44.9 & 78.1 & 45.5 & 38.1 & 55.7 & 74.9 & 82.7 & 51.4 & 61.2 \\ 
Rule-Base & 46.3 & 77.6 & 45.7 & 36.4 & 55.7 & 74.8 & 80.3 & 46.4 & 63.1 \\
\midrule \rowcolor{red4!20}
\multicolumn{10}{c}{Ours} \\
\abbr & 45.4 & 78.0 & 46.4 & 38.8 & 56.3 & \textbf{75.9} & 84.0 & \textbf{54.0} & 60.9 \\
\abbr+Rule-Base & \textbf{47.5} & \textbf{78.6} & \textbf{48.3} & \textbf{39.4} & \textbf{56.9} & 75.4 & \textbf{85.0} & \textbf{54.0} & \textbf{64.3} \\ 
\bottomrule
\end{tabular}%
}
\end{table}
\section{Conclusion and Limitation}

In this paper, we present a comprehensive ``recipe'' for building a high-performance MRM. Through extensive ablation studies, we systematically investigate every critical aspect of the development pipeline, including reward modeling paradigms, architectural design of the reward head, training regularization strategies, data curation, the choice of model backbone and scale, and ensemble methods. Our findings indicate that a simple yet optimized Naive-RM architecture—specifically, one with a two-layer MLP reward head using the SiLU activation function and trained without auxiliary regularization losses—is both efficient and highly effective. We demonstrate the critical importance of data curation, showing that a carefully selected blend of high-quality multimodal and text-only preference data is essential. Surprisingly, we found that text-only data can significantly enhance an MRM's judgment on multimodal tasks, particularly in dimensions like safety and mathematics. 

Based on these insights, we introduce \abbr{}, a powerful and efficient baseline for multimodal reward modeling. \abbr{} establishes a new state-of-the-art on several major MRM benchmarks, including MM-RLHF-Reward Bench and VL-Reward Bench, outperforming previous open-source and proprietary models. To demonstrate its practical utility, we integrate \abbr{} into a reinforcement learning pipeline, where it serves as an effective reward signal, consistently improving the performance of an MLLM across perception, reasoning, and conversational tasks. Despite our contributions, this work has certain limitations. First, due to computational resource constraints, we do not explore reward models based on backbones of 72B parameters or larger. Whether scaling up further would yield significant performance gains remains an open question. Second, our experiments show that for pure-text reward modeling tasks, LLM-based models currently outperform their MLLM-based counterparts. Whether a specific training strategy exists that could enable a multimodal model to surpass a comparable LLM-based reward model on pure-text benchmarks is still an open research problem.

\bibliography{colm2024_conference}
\bibliographystyle{unsrtnat}


\end{document}